\documentclass[table,11pt]{article}

\usepackage[]{EMNLP2023}

\usepackage{times}
\usepackage{latexsym}
\usepackage{multirow} 
\usepackage{booktabs}

\usepackage[T1]{fontenc}

\usepackage[utf8]{inputenc}

\usepackage{microtype}

\usepackage{inconsolata}
\usepackage{graphicx}

\usepackage{multirow}
\title{Expanding FLORES+ Benchmark for more Low-Resource Settings: Portuguese-Emakhuwa Machine Translation Evaluation}

\author{Felermino D. M. A. Ali\textsuperscript{1,2,3,5}, Henrique Lopes Cardoso\textsuperscript{1,2}, Rui Sousa-Silva\textsuperscript{3,4} \\
\textsuperscript{1}Laboratório de Inteligência Artificial e Ciência de Computadores (LIACC / LASI) \\
\textsuperscript{2}Faculdade de Engenharia da Universidade do Porto, Rua Dr. Roberto Frias, 4200-465 Porto, Portugal \\
\textsuperscript{3}Centro de Linguística da Universidade do Porto (CLUP) \\
\textsuperscript{4}Faculdade de Letras da Universidade do Porto, Via Panorâmica, 4150-564 Porto, Portugal \\
\textsuperscript{5}Faculdade de Engenharia da Universidade Lúrio, Pemba 3203, Mozambique \\
\texttt{\{up202100778, hlc\}@fe.up.pt, rssilva@letras.up.pt}
}

\begin{document}
\maketitle
\begin{abstract}

As part of the Open Language Data Initiative shared tasks, we have expanded the FLORES+ evaluation set to include Emakhuwa, a low-resource language widely spoken in Mozambique. We translated the \textit{dev} and \textit{devtest} sets from Portuguese into Emakhuwa, and we detail the translation process and quality assurance measures used. Our methodology involved various quality checks, including post-editing and adequacy assessments. The resulting datasets consist of multiple reference sentences for each source. 
We present baseline results from training a Neural Machine Translation system and fine-tuning existing multilingual translation models. Our findings suggest that spelling inconsistencies remain a challenge in Emakhuwa.  Additionally, the baseline models underperformed on this evaluation set, underscoring the necessity for further research to enhance machine translation quality for Emakhuwa.
The data is publicly available at \url{https://huggingface.co/datasets/LIACC/Emakhuwa-FLORES}

\end{abstract}

\section{Introduction}

Evaluation datasets and benchmarks are essential for advancing Natural Language Processing (NLP) models. They provide the necessary tools for assessing model performance and guiding further enhancements. However, the scarcity of evaluation datasets and benchmarks for low-resource languages has significantly hindered the progress of NLP technologies in these languages.
Recognizing this challenge, the FLORES+ evaluation set has emerged as a critical tool for the Machine Translation (MT) community, especially in low-resource languages. It promotes a more inclusive approach to language technology development across diverse linguistic landscapes.
This work focuses on expanding the FLORES+ evaluation set to include Emakhuwa, a low-resource language spoken in Mozambique by approximately 9 million people. Our dataset consists of the \textit{dev} and \textit{devtest} sets managed by the Open Language Data Initiative\footnote{https://oldi.org/}(OLDI), which contain 997 sentences and 1012 sentences, respectively. Throughout our data collection process, we implemented robust quality assurance mechanisms, including thorough post-editing. The resulting dataset features multiple reference translations derived from these post-editing efforts.

\section{Related Works}

The Flores v1.0 machine translation evaluation set was introduced by \citealp{guzman-etal-2019-flores}. This initial version focused on two language pairs: Nepali–English and Sinhala–English, with the data divided into \textit{dev}, \textit{test}, and \textit{devtest} splits.
After its release, the dataset was gradually expanded to include more languages. A significant expansion came with the work of \citealp{goyal2021flores101evaluationbenchmarklowresource}, who introduced Flores-101, extending the evaluation set to support 101 languages. Further expansion was done with the release of Flores-200 by the NLLB team \cite{nllb-22} in 2022, which increased the language coverage to 204 languages. Additional contributions include \citealp{doumbouya-etal-2023-machine}, who added the Nko language, and \citealp{indictrans2-23}, who incorporated Bodo, Dogri, Meitei, Sindhi, and Goan Konkani into the dataset. These contributions have significantly broadened the opportunities for low-resource languages in machine translation, allowing researchers to track the progress of MT systems on these expanded evaluations. However, the coverage remains limited, especially considering that there are over 7,000 languages worldwide. One such language that remains underserved is Emakhuwa, which still lacks datasets for machine translation.

\section{Emakhuwa}

Emakhuwa, alternatively referred to as Makua, Macua, or Makhuwa, belongs to the Bantu language family and is predominantly spoken in the northern and central regions of Mozambique, specifically in the Nampula, Niassa, Cabo Delgado, and Zambezia provinces.
There are eight variants of Emakhuwa, with Emakhuwa-Central (ISO 639-3 code \textit{vmw}) being the standard variety \cite{Ngunga_2014}. 

Emakhuwa follows the Subject-Verb-Object (SVO) structure, uses Latin scripts (ISO 15924 \textit{Latn}), and is gender-neutral. Furthermore, similarly to other languages in the Bantu family, it is linguistically rich, with complex morphology featuring agglutinative and tonal attributes.

\subsection{Challenges in Emakhuwa}


Emakhuwa digital resources are scarce, and the spelling standards are still under development. While a fully standardized system is not yet in place, the existing guidelines \cite{Ngunga_2014} offer a critical framework for contemporary written communication in Emakhuwa. One problem stressed in official standardization  \cite{Ngunga_2014} is the lack of guidance on tonal marking. Consequently, existing materials exhibit inconsistent spelling, particularly when marking tone, which is essential in Emakhuwa for disambiguation. 
To give an example, let us consider two words carrying distinct meanings: \textit{omala} and \textit{omaala} / \textit{omàla}; \textit{omala} means ``to finish", while \textit{omaala} / \textit{omàla} means ``to silence" or "to hush." In this case, the tonal marker \textit{aa} / \textit{à} clarifies the intended meaning. 

Spelling variations are largely evident in existing Emkahuwa text corpora, where some use diacritics (e.g., \textit{à}, \textit{è}, \textit{ì}, \textit{ò}, \textit{ù}) and consonantal sounds (e.g, \textit{kh}, \textit{nn}) for tonal marking, others use vowel lengthening (e.g., \textit{aa}, \textit{ee}, \textit{ii}, \textit{oo}, \textit{uu}), and some even use a combination of methods. Emakhuwa's agglutinative nature with complex morphology further amplifies spelling discrepancies. Since tonal variations often occur at the morpheme level, different combinations of morphemes result in varied spellings of the same word.

These spelling inconsistencies create significant obstacles for language technology processes. They lead to data sparsity, as some spelling variants appear less frequently, which impairs the model’s ability to learn the language’s nuances effectively. This sparsity inflates the vocabulary size and can result in reduced performance of language technologies.

An additional challenge in Emakhuwa that contributes to inconsistencies is the adaptation of loanwords. Emakhuwa text corpora frequently contain Portuguese loanwords with inconsistent adaptations due to the absence of standardized guidelines for integrating borrowed terms~\cite{ali-etal-2024-detecting}. These loanwords are adapted in one of three ways: phonetically to match Portuguese pronunciation, in alignment with Emakhuwa phonotactics, or retained unchanged from Portuguese.




\section{Methodology}

We chose to translate the \textit{devtest} and \textit{dev} sets from Portuguese (\textit{pt}) into Emakhuwa (\textit{vmw}) because our translators were only proficient in these two languages. We focus specifically on the central variant of Emakhuwa, as it is the standard and established language variant.

The translators were selected based on their proficiency in these languages and their proven experience in Portuguese-Emakhuwa translation. In total, we collaborated with five experts: two were assigned the tasks of translation and revision, while the remaining collaborators were responsible for evaluating the translations (refer to Table~\ref{tab:team} in Appendix for more details). 

\begin{figure*}
  \centering
  \includegraphics[width=1\textwidth]{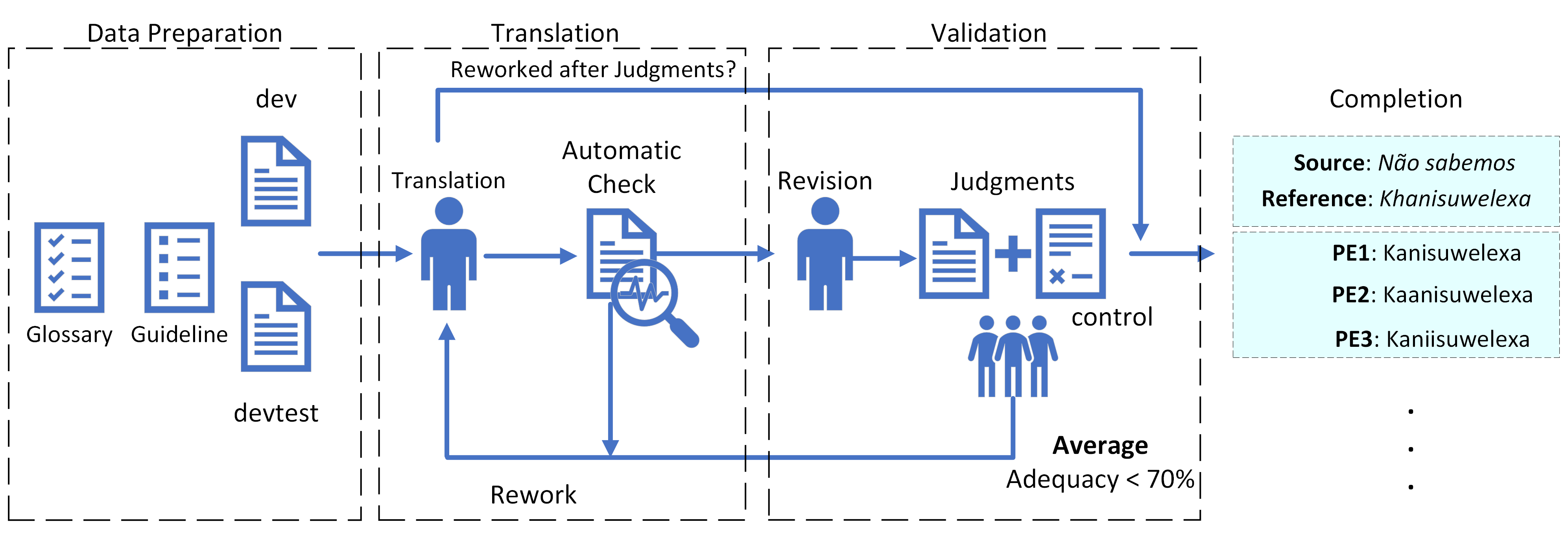}
  \caption{Workflow}
  \label{fig:translation-workflow}
\end{figure*}

In general, we implemented the workflow as the peer review process, divided into three main steps: Data Preparation, Translation, and Validation. Below is a detailed description of each step (refer to Figure~\ref{fig:translation-workflow}).

\subsection{Data Preparation}

We compile the sentences in \textit{devtest} and \textit{dev} sets as segments and then load them to the Matecat\footnote{https://www.matecat.com/} CAT (Computer-Assisted Translation) tool. Before assigning translation tasks, we prepare a guideline and glossary. The guidelines were adapted from the Open Language Data Initiative guidelines\footnote{https://oldi.org/guidelines}, written in Portuguese and suggesting that the translated text should adhere to the latest orthography standards of the central variant of Emakhuwa.
On the other hand, the glossary was built by digitizing existing bilingual dictionaries and the glossary of Political, Sports, and Social Concepts from Radio of Mozambique~\cite{mozambique2016glossarios}. We conducted a small workshop to familiarize the team with the guidelines and gather feedback to improve them. The translation team found the glossary helpful, as it prevented using loanwords for existing Emakhuwa terms and ensured consistency in translations. 

\subsection{Translation}

Translation tasks were subdivided between two translators: one worked on the \textit{devtest} segments, and the other on the \textit{dev} segments. Once all segments were translated, they were submitted to our spell checker system for an automatic check to identify potential misspellings (refer to Figure~\ref{fig:spelling-report} in the Appendix). We then provided feedback to the translators, asking them to review and refine their work if necessary. 

\subsection{Validation}

The validation corresponds to two steps: revision and Judgments.

\subsubsection{Revision} 

Following the translation step, we swapped the translated works between the two translators, asking them to post-edit each other's translations on the Matecat platform. Table~\ref{tab:revison_statitics} provides the Quality Report generated by Matecat, which includes various metrics used to evaluate the translation based on the revisions made. The report indicates that the reviewer working on \textit{devtest} made more suggestions. A closer examination of the error categories on \textit{devtest} (refer to Figure~\ref{fig:errors-points} in the Appendix) reveals that most of the issues identified in the translation fell under the category of "Language Quality", meaning grammar, punctuation, and spelling errors. On the other hand, the reviewer of the \textit{dev} set identified mostly errors related to "terminology and language consistency", suggesting that the translator was not consistently using the proper terms and maintaining uniformity throughout the text.

\begin{table}[h]
\centering
\resizebox{0.45\textwidth}{!}{
\begin{tabular}{rcc}
\toprule
 & \textbf{dev} & \textbf{devtest} \\
\midrule
Post-Editing Effort & 99\% & 95\%\\
Time to edit & 02m38s & 05m42s\\
Quality score & 23.31 & 54.22\\
Avg. Edit Distance & 0.23 $\pm$ 1.77 & 7.09 $\pm$ 11.94 \\
\bottomrule
\end{tabular}
}
\caption{Matecat's quality report post revision. }
\label{tab:revison_statitics}
\end{table}

\subsubsection {Judgments} 
Once all segments have been revised, we perform a second translation quality assessment using a Direct Assessment (DA) pipeline similar to the one described by \citealt{guzman-etal-2019-flores}. Judgments were collected using our annotation tool (see Figure~\ref{fig:annotation-tool} in the Appendix), and involve the following aspects.

\paragraph{Direct Assessment} Three different raters evaluate the translation adequacy (i.e., the perceived translation quality) on a scale from 0 to 100. A score of 0 means that "no meaning was preserved in the translation". Scores from 1 to 34 - "the translation preserves some of the source meaning but loses significant parts", 35 and 67 - "the translation retains most of the source meaning", 68 to 99 - "the translation is consistent with the source text", and a score of 100 means "the translation is perfect". These quality intervals are inspired by the study of~\citealt{wang-etal-2024-afrimte}.

\paragraph{Control} To ensure raters' attentiveness and improve consistency during the evaluation, we included control instances with incorrect translation pairs. These incorrect pairs were generated using the Madland-400-3bt\footnote{https://huggingface.co/google/madlad400-3b-mt} model \cite{MADLAD-400-Kudugunta}, a multilingual MT system that supports the Emetto variant of Emakhuwa (ISO 639-3 \textit{mgh}). While this model typically performs poorly when translating from Portuguese to Emakhuwa, it produces similar words that can mislead inattentive annotators.
Based on these control translations, we provided feedback to the evaluators as they progressed in their tasks. We used emojis to give the feedback in our annotation tool: a \includegraphics[height=1em]{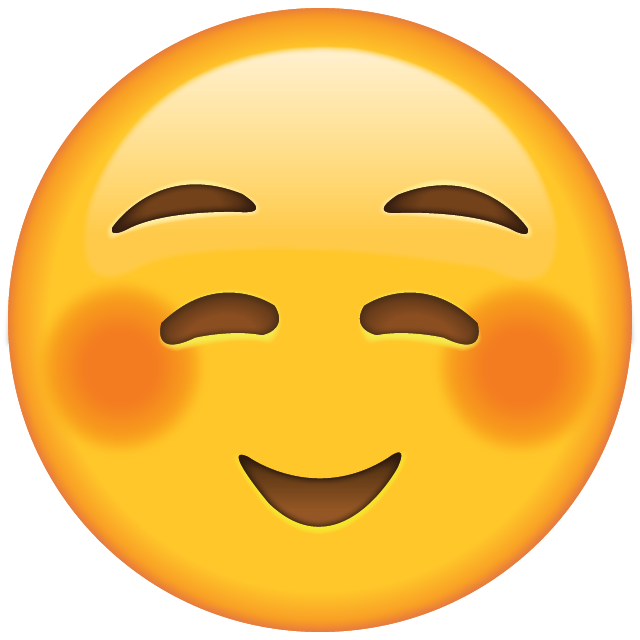} appeared if less than 25\% of control translations were incorrectly rated (i.e. scores above 34 points), \includegraphics[height=1em]{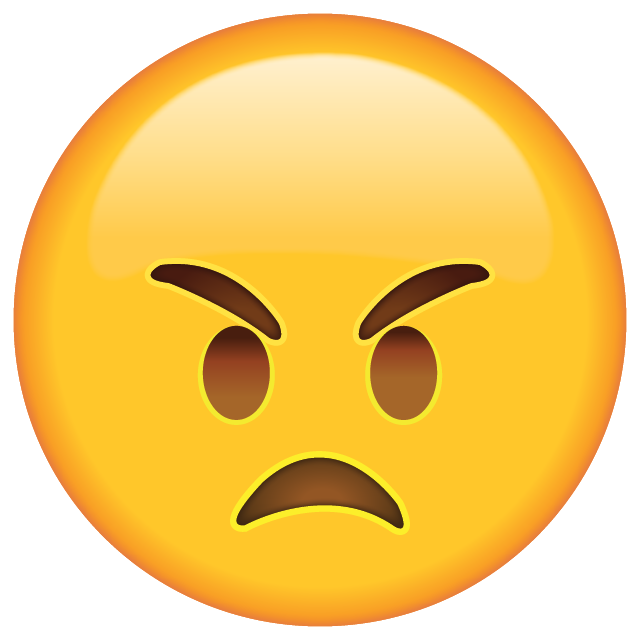} if 25\%-50\% are incorrectly rated, \includegraphics[height=1em]{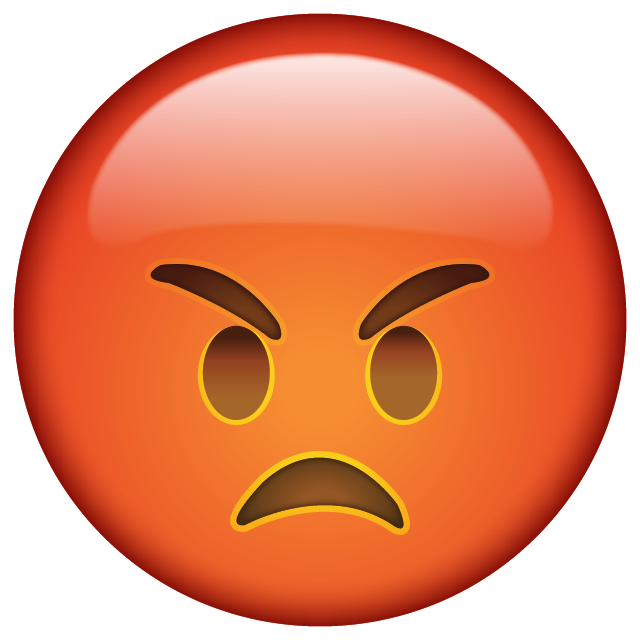} if 50\%-75\% are incorrectly rated, and \includegraphics[height=1em]{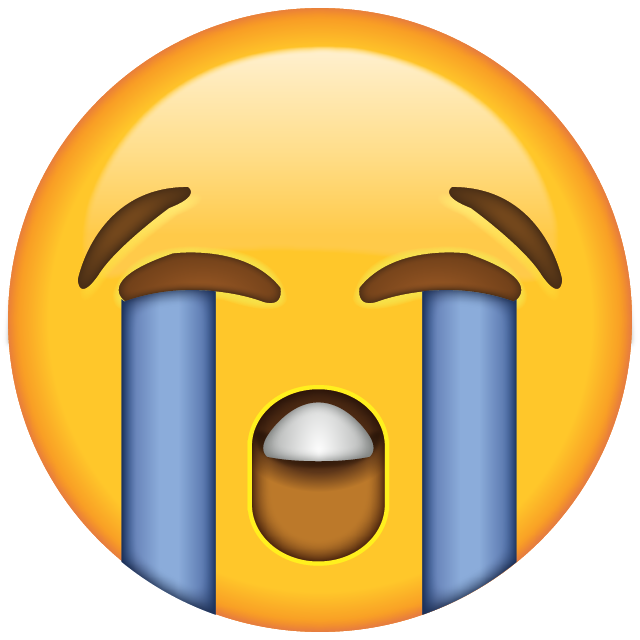} if more than 75\% of control translations are rated too highly.

\paragraph{Post Editing}
During the validation phase, we asked evaluators to post-edit translations with lower scores to enhance fluency and better align them with the source sentence's meaning. However, this task was made optional to prevent evaluators from inflating scores to avoid additional post-editing work.

\paragraph{Standard orthography} To assess the perceived usage of standard orthography, raters also judged whether the translated text used standard orthography on a scale from 1 (not using standard orthography) to 5 (entirely written in standard orthography).

\begin{figure}[h]
  \centering
  \includegraphics[width=0.5\textwidth]{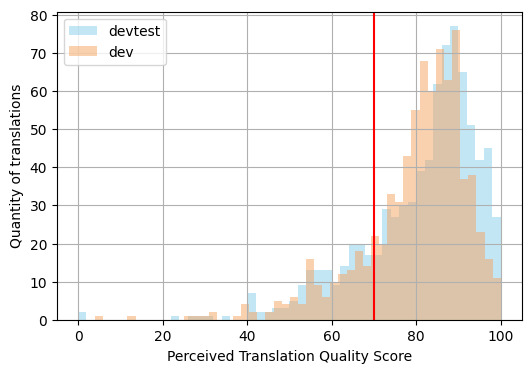}
  \caption{Averaged Translation Quality Score Histogram on both \textit{dev} and \textit{devtest} sets. Translations with an average score below 70 (indicated by the red line) were returned to the translator for rework.}
  \label{fig:histogram}
\end{figure}

Finally, we calculate the average score for each segment. We then returned segments scoring below 70 to the translator for rework. Figure~\ref{fig:histogram} shows the histogram of the average translation scores.

\subsection{Analysis}

\begin{figure}[h]
  \centering
  \includegraphics[width=0.5\textwidth]{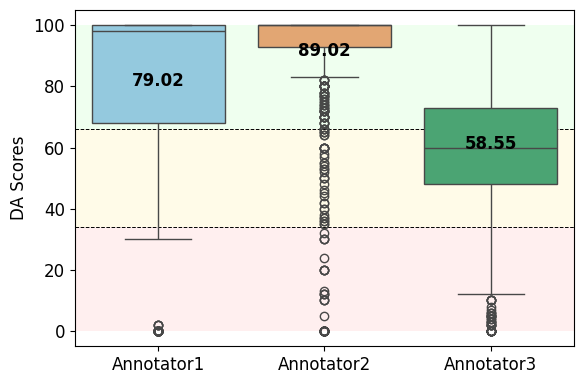}
  \caption{Direct assessment adequacy scores per Annotator on \textit{dev} set}
  \label{fig:da_manual_dev}
\end{figure}

\begin{figure}[h]
  \centering
  \includegraphics[width=0.5\textwidth]{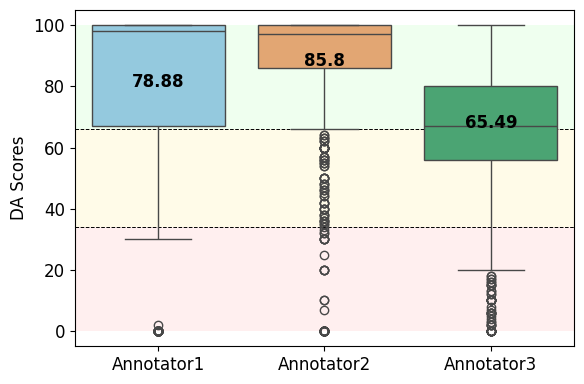}
\caption{Direct assessment adequacy scores per Annotator on \textit{devtest} set}
  \label{fig:da_manual_devset}
\end{figure}

\begin{figure}[h]
  \centering
  \includegraphics[width=0.5\textwidth]{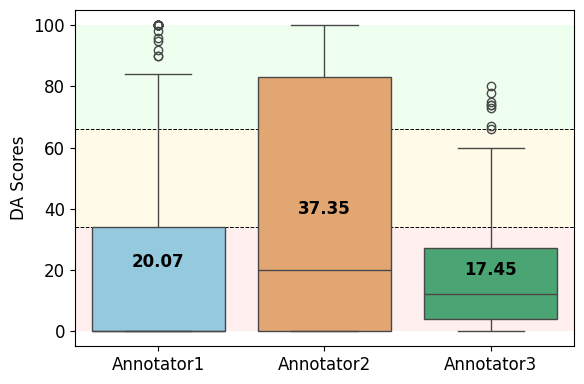}
  \caption{Direct assessment adequacy scores per Annotator on \textit{control} set}
  \label{fig:da_control}
\end{figure}

Figures~\ref{fig:da_manual_dev} and~\ref{fig:da_manual_devset} show the raw scores per annotator for direct assessments. Given the mean scores, in both the \textit{test} and \textit{devtest} sets, Annotator 1 and Annotator 2 gave higher quality scores, while Annotator 3 was more critical but still within the spectrum of acceptable translations. This suggests a generally positive perception of the translations produced.
Figure~\ref{fig:da_control} displays the direct Assessment scores on the control set. Annotator 1 and Annotator 3 have median scores below the threshold of 34 points, suggesting that, as expected, they have generally assessed the control translations as low quality. Annotator 2, however, has a median score above the threshold, suggesting a trend to a more positive assessment compared to the other two annotators and was less attentive among the annotator. 

\begin{figure}[h]
  \centering
  \includegraphics[width=0.5\textwidth]{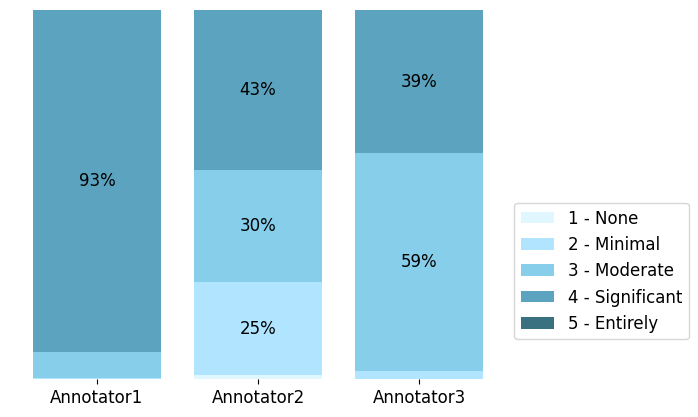}
    \caption{Assessment of standard orthography usage on the control set.}
  \label{fig:ortho_control}
\end{figure}

Table~\ref{tab:icc_results_combined} provides the reliability results for adequacy and standard orthography usage assessments. The inter-class correlation for adequacy is 0.67 for \textit{dev} and 0.66 for \textit{devtest}, suggesting moderate reliability. However, the inter-class correlations for standard orthography usage are lower, with values of 0.35 for \textit{dev} and 0.27 for \textit{devtest}, indicating considerable disagreement among annotators. This discrepancy highlights the ongoing lack of clarity regarding Emakhuwa spelling standards, as further illustrated in Figure~\ref{fig:ortho_control}, which depicts the varying assessments of standard orthography.

\begin{table}[h]
\centering
\resizebox{0.5\textwidth}{!}{%
\begin{tabular}{rcccc}
\toprule
  & \multicolumn{2}{c}{\textbf{Adequacy}}  & \multicolumn{2}{c}{\textbf{Orthography}}  \\ \cmidrule(rl){2-3} \cmidrule(rl){4-5}
  & dev & devtest & dev & devtest \\
\midrule
ICC & 0.67 & 0.66 & 0.35 & 0.27 \\
CI & [0.63, 0.71] & [0.62, 0.7] & [0.27, 0.42] & [0.18, 0.35]\\ 
\bottomrule
\end{tabular}%
}
\caption{Intraclass Correlation Coefficient (ICC) and Confidence Interval (CI) Results for Adequacy and Orthography usage annotation. }
\label{tab:icc_results_combined}
\end{table}

\subsection{Dataset Collected}

Table~\ref{tab:data_statitics} presents the statistics for the \textit{devtest} and \textit{dev} sets resulting from the completion of the translation tasks. The \textit{dev} set comprises 997 sentence pairs, while the \textit{devtest} set contains 1,012 sentence pairs. A sample of the dataset is displayed in Table~\ref{tab:translations} in the Appendix.

\begin{table}
\centering
\resizebox{0.4\textwidth}{!}{
\begin{tabular}{rcc}
\toprule
 & \textbf{dev} & \textbf{devtest} \\
\midrule
Refs. & 997 & 1,012\\
Refs. words & 18,673 & 21,011 \\
Refs. post-edits & 1,848 & 1,889 \\
\bottomrule
\end{tabular}
}
\caption{Statitics for the resulting dataset sets}
\label{tab:data_statitics}
\end{table}

\section{Experiments}\label{sec:Expermient}

This section describes the experiment involving training Neural MT models using the training sets described below. Then, we performed a comprehensive benchmark evaluation using the evaluation sets introduced in this study.

\subsection{Training Data}

To train the models, we used the data outlined below:
\begin{itemize}

\item {~\citealp{Felermino2021} dataset:} This subset comprises parallel data in Portuguese and Emakhuwa from different sources, including online texts from the Jehovah's Witness, the African Story Book websites, and Optical Character Recognition (OCR) extracted texts. The corpus contains diverse writing styles, spelling styles, and genres.
\item {Parallel News:} This subset consists of news articles translated from Portuguese into Emakhuwa.

\end{itemize}

The dataset includes around 63k training parallel sentences and 964 validation parallel sentences, spanning a range of topics (see Table~\ref{tab:training-data-stats}), where a significant portion of the data comes from the religious domain, mainly consisting of translations of biblical texts. 

\begin{table}[h]
\centering

\resizebox{0.5\textwidth}{!}{%
\begin{tabular}{rcccc}
\toprule
 &  \multicolumn{2}{c}{\textbf{Sentences}} & \multicolumn{2}{c}{\textbf{Tokens}} \\ \cmidrule(lr){2-3} \cmidrule(lr){4-5}
 \textbf{Source} & Train & Dev & \textit{pt} & \textit{vmw}\\ 
\midrule

\citealp{Felermino2021} & 46,454 & 399 & 1,104,279 & 951,520 \\
News    & 17,403 & 565 & 596,066 & 541,598\\
\midrule
\textbf{Total}   & 63,857 & 964 & 1,700,345 & 1,493,118 \\
\bottomrule
\end{tabular}%
}
\caption{Training and Validation data statistics}
\label{tab:training-data-stats}
\end{table}

\begin{table*}[h]
\centering

\resizebox{1\textwidth}{!}{
\begin{tabular}{cl|cc|cc|cc|cc}
\toprule
    & & \multicolumn{4}{c}{\textit{\textbf{dev}}} & \multicolumn{4}{c}{\textit{\textbf{devtest}}} \\ \cmidrule(rl){3-6} \cmidrule(rl){7-10}
    & & \multicolumn{2}{c}{\textbf{Single Ref.}} & \multicolumn{2}{c}{\textbf{Multi Ref.}} & \multicolumn{2}{c}{\textbf{Single Ref.}} & \multicolumn{2}{c}{\textbf{Multi Ref.}} \\ \cmidrule(rl){3-4} \cmidrule(rl){5-6} \cmidrule(rl){7-8} \cmidrule(rl){8-10}
& & BLEU & chrF & BLEU & chrF & BLEU & chrF & BLEU & chrF    \\
\toprule

\multicolumn{10}{l}{\textbf{Small Transformer}} \\ 
\hline
\multirow{2}{*}{\textbf{Baseline}} & pt $\to$ vmw & 3.7 & 30.67 & 3.95$_{(+0.25)}$ & 31.32$_{(+0.65)}$ & 3.27 & 29.23 & 3.57$_{(+0.3)}$ & 29.84$_{(+0.61)}$   \\ 
&  vmw $\to$ pt & 4.36 & 25.48 & - & - & 2.93 & 23.96  & - & -  \\
\hline
\multicolumn{10}{l}{\textbf{Multilingual Language Models}} \\ 
\hline

\multirow{2}{*}{\textbf{afri-byT5}} & pt $\to$ vmw & 10.32 & 41.88 & 10.81$_{(+0.49)}$ & 42.64$_{(+0.76)}$ & 7.03 & 35.87 & 7.73$_{(+0.7)}$ & 36.72$_{(+0.85)}$  \\ 
&  vmw $\to$ pt & \textbf{22.45} & \textbf{47.31} & - & - & 13.74 & \textbf{37.78} & - & -  \\ 
\multirow{2}{*}{\textbf{afri-mT5}} &  pt $\to$ vmw & 5.66 & 35.37 & 5.96$_{(+0.3)}$ & 36.01$_{(+0.64)}$ & 4.7 & 32.7 & 5.06$_{(+0.36)}$ & 33.25$_{(+0.55)}$   \\ 
&  vmw $\to$ pt & 12.12 & 38.18 & - & - & 7.39 & 32.92 & - & -  \\
\multirow{2}{*}{\textbf{ByT5}}  &  pt $\to$ vmw & \textbf{10.66} & \textbf{42.37} & \textbf{11.2}$_{(+0.54)}$ & \textbf{43.16}$_{(+0.79)}$ & \textbf{7.49} & \textbf{36.33} & \textbf{8.13}$_{(+0.64)}$ & \textbf{37.15}$_{(+0.82)}$ \\
&  vmw $\to$ pt & 22.24 & 47.01 & - & - & \textbf{14.1} & 37.75 & - & -   \\
\multirow{2}{*}{\textbf{MT0}}  &   pt $\to$ vmw & 5.52 & 30.33 & 5.76$_{(+0.24)}$ & 30.9$_{(+0.57)}$ & 4.69 & 27.89 & 5.02$_{(+0.33)}$ & 28.36$_{(+0.47)}$   \\ 
&  vmw $\to$ pt & 17.46 & 38.92 & - & - & 10.63 & 32.69 & - & -  \\ 
\multirow{2}{*}{\textbf{mT5}}  &  pt $\to$ vmw & 6.76 & 34.09 & 7.18$_{(+0.42)}$ & 34.8$_{(+0.71)}$ & 5.67 & 31.67 & 6.06$_{(+0.39)}$ & 32.23$_{(+0.56)}$   \\ 
&  vmw $\to$ pt & 15.42 & 37.58 & - & - & 9.65 & 32.22 & - & - \\ 
 \hline
\multicolumn{10}{l}{\textbf{Many-to-Many Multilingual Translation Language Models}} \\ 
\hline
\multirow{2}{*}{\textbf{M2M100}} &  pt $\to$ vmw & 8.25 & 39.22 & 8.79$_{(+0.54)}$ & 40.14$_{(+0.92)}$  & 6.92 & 36.33 & 7.57$_{(+0.65)}$ & 37.19$_{(+0.86)}$ \\ 
&  vmw $\to$ pt & 21.08 & 45.31 & - & - & 13.67 & 37.46 & - & - \\ 
\multirow{2}{*}{\textbf{NLLB}}  &  pt  $\to$ vmw & 8.19 & 41.44 & 8.74$_{(+0.54)}$ & 42.32$_{(+0.88)}$ & 5.88 & 36.13 & 6.34$_{(+0.46)}$ & 37.01$_{(+0.88)}$ \\ 
&  vmw $\to$ pt & 17.41 & 42.88 & - & - & 10.35 & 35.05 & - & - \\ 

\bottomrule
\end{tabular}
}
\caption{BLEU and chrF scores for various models on \textit{dev} and \textit{devtest} splits, for single and multiple references}
\label{tab:results-scores}
\end{table*}

\subsection{Setup}\label{sec:setup}

We trained MT models in both directions, \textit{pt-vmw} (Portuguese to Emakhuwa) and \textit{vmw-pt} (Emakhuwa to Portuguese), using two approaches: training a vanilla transformer model and fine-tuning existing multilingual language models.

\paragraph{Training} 
We adopt transformer architecture~\cite{Vaswani2017}, implemented through the OpenNMT toolkit \cite{klein-etal-2017-opennmt}. The model consists of an encoder and decoder comprising 6 layers, 8 heads, and 512 hidden units in the feed-forward network. We used an embedding size of 512 dimensions for both source and target words and a batch size of 32.  
We applied layer normalization and added dropout with a 0.1 probability to the embedding and transformer layers. Additionally, the Adam optimizer \cite{Kingma2014AdamAM} was used, and a learning rate of 0.0002. The checkpoints were saved every 1000 updates. We preprocess the input, applying the Byte Pair Encoding subword segmentation.

\paragraph{Fine-tuning Multilingual Models} 
Multilingual language models are one of the most prominent approaches to low-resource languages nowadays. Since it enables knowledge transfer among related languages, making cross-lingual transfer and zero-shot learning possible. 

In our experiments, we fine-tuned various multilingual language models that are well-established in the literature, namely: mT5 \cite{xue-etal-2021-mt5}, ByT5 \cite{xue-etal-2022-byt5}, and the multilingual translation models M2M-100~\cite{Angela-Fan-m2m} and NLLB~\cite{Costa-jussa2024}. Specifically, we use MT5-base (580M parameters), ByT5-base (580M parameters), M2M-100 (418M parameters), and NLLB-200's distilled variant (600M parameters).
Additionally, we also fine-tuned the African-centric language models, namely, AfriByT5 (580M parameters) and AfriMT5 (580M parameters) by \citeauthor{adelani-etal-2022-thousand}, \citeyear{adelani-etal-2022-thousand}.  

\subsection{Evaluation}
To assess the systems' performance, we used the SacreBLEU \cite{post-2018-call} to compute the BLEU~\cite{papineni-etal-2002-bleu} and ChrF scores~\cite{popovic-2015-chrf}.

\section{Results and Discussion}

Results are presented in Table~\ref{tab:results-scores}. Our baseline results, derived from a small Transformer model, set a foundational performance benchmark. On the \textit{devtest} set, the baseline model achieved a BLEU score of 2.93 and a ChrF score of 23.96 for the \textit{vmw} $\to$ \textit{pt} translation direction. These modest scores underscore the limitations of the small Transformer model in handling the complexities of translation tasks involving low-resource languages like Emakhuwa.

However, introducing multilingual language models enhanced translation performance, particularly in the \textit{vmw} $\to$ \textit{pt} direction. Among these, models based on ByT5 demonstrated superior performance. For instance, the fine-tuned ByT5 model achieved a BLEU score of 14.1 and a ChrF score of 37.75 on the \textit{devtest} set, which marks a substantial improvement over the baseline. This highlights the advantage of leveraging tokenization-free approaches, which are better suited for handling the morphological richness and orthographic variations characteristic of Emakhuwa.

Across Table~\ref{tab:results-scores}, our results show that while BLEU scores remained relatively low in the \textit{pt} $\to$ \textit{vmw} translation direction, ChrF were consistently higher. This discrepancy between BLEU and ChrF scores suggests that BLEU may be disproportionately penalizing spelling variations and minor orthographic differences, which are more prevalent in Emakhuwa translations. ChrF, on the other hand, being more sensitive to character-level n-grams, captures better the quality of translations. Nevertheless, further studies need to be done to assess the correlation of these automatic metrics with human evaluations.

\paragraph{Using multiple references}  Notably, using multiple references improved scores for both BLEU and ChrF across all models. Specifically, BLEU scores increased by +0.24 to +0.54 on the \textit{dev} set and by +0.3 on the \textit{devtest} set.

\section{Conclusion}
In conclusion, this study expanded the FLORES+ evaluation set to include Emakhuwa, a low-resource language spoken in Mozambique. By translating the \textit{dev} and \textit{devtest} sets from Portuguese to Emakhuwa. We discussed key challenges such as spelling inconsistencies and loanword adaptations, which are prevalent due to Emakhuwa's underdeveloped spelling standards. Our rigorous methodology, involving translation, post-editing, and validation, ensured high-quality datasets used to benchmark Neural MT models. The results indicate that incorporating multiple reference translations can enhance translation quality, particularly in languages with underdeveloped orthographies such as Emakhuwa. The dataset is publicly available, providing a valuable resource for future research in low-resource language MT.





\section*{Acknowledgements}
This work was financially supported by Base Funding (UIDB/00027/2020) and Programmatic Funding (UIDP/00027/2020) of the Artificial Intelligence and Computer Science Laboratory (LIACC) funded by national funds through FCT/MCTES (PIDDAC) as well as supported by the Base (UIDB/00022/2020) and Programmatic (UIDP/00022/2020) projects of the Centre for Linguistics of the University of Porto.
Felermino Ali is supported by a PhD studentship (with reference SFRH/BD/151435/2021), funded by Funda\c{c}\~{a}o para a Ci\^{e}ncia e a Tecnologia (FCT).

We sincerely thank the Lacuna Fund for their generous sponsorship, which made the creation of this dataset possible. Our gratitude also goes to the Translation Team for their dedication and hard work on this project.


\bibliographystyle{acl_natbib}
\bibliography{anthology}

\begin{thebibliography}{22}
\expandafter\ifx\csname natexlab\endcsname\relax\def\natexlab#1{#1}\fi

\bibitem[{Adelani et~al.(2022)Adelani, Alabi, Fan, Kreutzer, Shen, Reid, Ruiter, Klakow, Nabende, Chang, Gwadabe, Sackey, Dossou, Emezue, Leong, Beukman, Muhammad, Jarso, Yousuf, Niyongabo~Rubungo, Hacheme, Wairagala, Nasir, Ajibade, Ajayi, Gitau, Abbott, Ahmed, Ochieng, Aremu, Ogayo, Mukiibi, Ouoba~Kabore, Kalipe, Mbaye, Tapo, Memdjokam~Koagne, Munkoh-Buabeng, Wagner, Abdulmumin, Awokoya, Buzaaba, Sibanda, Bukula, and Manthalu}]{adelani-etal-2022-thousand}
David Adelani, Jesujoba Alabi, Angela Fan, Julia Kreutzer, Xiaoyu Shen, Machel Reid, Dana Ruiter, Dietrich Klakow, Peter Nabende, Ernie Chang, Tajuddeen Gwadabe, Freshia Sackey, Bonaventure F.~P. Dossou, Chris Emezue, Colin Leong, Michael Beukman, Shamsuddeen Muhammad, Guyo Jarso, Oreen Yousuf, Andre Niyongabo~Rubungo, Gilles Hacheme, Eric~Peter Wairagala, Muhammad~Umair Nasir, Benjamin Ajibade, Tunde Ajayi, Yvonne Gitau, Jade Abbott, Mohamed Ahmed, Millicent Ochieng, Anuoluwapo Aremu, Perez Ogayo, Jonathan Mukiibi, Fatoumata Ouoba~Kabore, Godson Kalipe, Derguene Mbaye, Allahsera~Auguste Tapo, Victoire Memdjokam~Koagne, Edwin Munkoh-Buabeng, Valencia Wagner, Idris Abdulmumin, Ayodele Awokoya, Happy Buzaaba, Blessing Sibanda, Andiswa Bukula, and Sam Manthalu. 2022.
\newblock \href {https://doi.org/10.18653/v1/2022.naacl-main.223} {A few thousand translations go a long way! leveraging pre-trained models for {A}frican news translation}.
\newblock In \emph{Proceedings of the 2022 Conference of the North American Chapter of the Association for Computational Linguistics: Human Language Technologies}, pages 3053--3070, Seattle, United States. Association for Computational Linguistics.

\bibitem[{AI4Bharat et~al.(2023)AI4Bharat, Gala, Chitale, AK, Doddapaneni, Gumma, Kumar, Nawale, Sujatha, Puduppully, Raghavan, Kumar, Khapra, Dabre, and Kunchukuttan}]{indictrans2-23}
AI4Bharat, Jay Gala, Pranjal~A. Chitale, Raghavan AK, Sumanth Doddapaneni, Varun Gumma, Aswanth Kumar, Janki Nawale, Anupama Sujatha, Ratish Puduppully, Vivek Raghavan, Pratyush Kumar, Mitesh~M. Khapra, Raj Dabre, and Anoop Kunchukuttan. 2023.
\newblock \href {http://arxiv.org/abs/arXiv:2305.16307} {Indictrans2: Towards high-quality and accessible machine translation models for all 22 scheduled indian languages}.

\bibitem[{Ali et~al.(2021)Ali, Caines, and Malavi}]{Felermino2021}
Felermino D. M.~A. Ali, Andrew Caines, and Jaimito L.~A. Malavi. 2021.
\newblock \href {https://doi.org/10.48550/ARXIV.2104.05753} {Towards a parallel corpus of portuguese and the bantu language emakhuwa of mozambique}.

\bibitem[{Ali et~al.(2024)Ali, Lopes~Cardoso, and Sousa-Silva}]{ali-etal-2024-detecting}
Felermino Dario~Mario Ali, Henrique Lopes~Cardoso, and Rui Sousa-Silva. 2024.
\newblock \href {https://aclanthology.org/2024.lrec-main.425} {Detecting loanwords in emakhuwa: An extremely low-resource {B}antu language exhibiting significant borrowing from {P}ortuguese}.
\newblock In \emph{Proceedings of the 2024 Joint International Conference on Computational Linguistics, Language Resources and Evaluation (LREC-COLING 2024)}, pages 4750--4759, Torino, Italia. ELRA and ICCL.

\bibitem[{Doumbouya et~al.(2023)Doumbouya, Dian{\'e}, Ciss{\'e}, Dian{\'e}, Sow, Doumbouya, Bangoura, Bayo, Conde, Dian{\'e}, Piech, and Manning}]{doumbouya-etal-2023-machine}
Moussa Doumbouya, Baba~Mamadi Dian{\'e}, Solo~Farabado Ciss{\'e}, Djibrila Dian{\'e}, Abdoulaye Sow, S{\'e}r{\'e}~Moussa Doumbouya, Daouda Bangoura, Fod{\'e}~Moriba Bayo, Ibrahima~Sory Conde, Kalo~Mory Dian{\'e}, Chris Piech, and Christopher Manning. 2023.
\newblock \href {https://doi.org/10.18653/v1/2023.wmt-1.34} {Machine translation for nko: Tools, corpora, and baseline results}.
\newblock In \emph{Proceedings of the Eighth Conference on Machine Translation}, pages 312--343, Singapore. Association for Computational Linguistics.

\bibitem[{Fan et~al.(2021)Fan, Bhosale, Schwenk, Ma, El-Kishky, Goyal, Baines, Celebi, Wenzek, Chaudhary, Goyal, Birch, Liptchinsky, Edunov, Auli, and Joulin}]{Angela-Fan-m2m}
Angela Fan, Shruti Bhosale, Holger Schwenk, Zhiyi Ma, Ahmed El-Kishky, Siddharth Goyal, Mandeep Baines, Onur Celebi, Guillaume Wenzek, Vishrav Chaudhary, Naman Goyal, Tom Birch, Vitaliy Liptchinsky, Sergey Edunov, Michael Auli, and Armand Joulin. 2021.
\newblock \href {http://jmlr.org/papers/v22/20-1307.html} {Beyond english-centric multilingual machine translation}.
\newblock \emph{Journal of Machine Learning Research}, 22(107):1--48.

\bibitem[{Goyal et~al.(2021)Goyal, Gao, Chaudhary, Chen, Wenzek, Ju, Krishnan, Ranzato, Guzman, and Fan}]{goyal2021flores101evaluationbenchmarklowresource}
Naman Goyal, Cynthia Gao, Vishrav Chaudhary, Peng-Jen Chen, Guillaume Wenzek, Da~Ju, Sanjana Krishnan, Marc'Aurelio Ranzato, Francisco Guzman, and Angela Fan. 2021.
\newblock \href {http://arxiv.org/abs/2106.03193} {The flores-101 evaluation benchmark for low-resource and multilingual machine translation}.

\bibitem[{Guzm{\'a}n et~al.(2019)Guzm{\'a}n, Chen, Ott, Pino, Lample, Koehn, Chaudhary, and Ranzato}]{guzman-etal-2019-flores}
Francisco Guzm{\'a}n, Peng-Jen Chen, Myle Ott, Juan Pino, Guillaume Lample, Philipp Koehn, Vishrav Chaudhary, and Marc{'}Aurelio Ranzato. 2019.
\newblock \href {https://doi.org/10.18653/v1/D19-1632} {The {FLORES} evaluation datasets for low-resource machine translation: {N}epali{--}{E}nglish and {S}inhala{--}{E}nglish}.
\newblock In \emph{Proceedings of the 2019 Conference on Empirical Methods in Natural Language Processing and the 9th International Joint Conference on Natural Language Processing (EMNLP-IJCNLP)}, pages 6098--6111, Hong Kong, China. Association for Computational Linguistics.

\bibitem[{Kingma and Ba(2014)}]{Kingma2014AdamAM}
Diederik~P. Kingma and Jimmy Ba. 2014.
\newblock \href {https://api.semanticscholar.org/CorpusID:6628106} {Adam: A method for stochastic optimization}.
\newblock \emph{CoRR}, abs/1412.6980.

\bibitem[{Klein et~al.(2017)Klein, Kim, Deng, Senellart, and Rush}]{klein-etal-2017-opennmt}
Guillaume Klein, Yoon Kim, Yuntian Deng, Jean Senellart, and Alexander Rush. 2017.
\newblock \href {https://www.aclweb.org/anthology/P17-4012} {{O}pen{NMT}: Open-source toolkit for neural machine translation}.
\newblock In \emph{Proceedings of {ACL} 2017, System Demonstrations}, pages 67--72, Vancouver, Canada. Association for Computational Linguistics.

\bibitem[{Kudugunta et~al.(2024)Kudugunta, Caswell, Zhang, Garcia, Xin, Kusupati, Stella, Bapna, and Firat}]{MADLAD-400-Kudugunta}
Sneha Kudugunta, Isaac Caswell, Biao Zhang, Xavier Garcia, Derrick Xin, Aditya Kusupati, Romi Stella, Ankur Bapna, and Orhan Firat. 2024.
\newblock Madlad-400: a multilingual and document-level large audited dataset.
\newblock In \emph{Proceedings of the 37th International Conference on Neural Information Processing Systems}, NIPS '23, Red Hook, NY, USA. Curran Associates Inc.

\bibitem[{Moçambique~E.P.(2016)}]{mozambique2016glossarios}
R.~de Moçambique~E.P. 2016.
\newblock \href {http://197.249.65.29/moodle/file.php/1/Glosario\_RMe.pdf} {Glossários de conceitos políticos, desportivos e sociais (português-línguas moçambicanas)}.
\newblock Retrieved from \url{http://197.249.65.29/moodle/file.php/1/Glosario\_RMe.pdf}.

\bibitem[{Ngunga and Faquir(2014)}]{Ngunga_2014}
Armindo Ngunga and Osvaldo Faquir. 2014.
\newblock \emph{Padroniza{\c{c}}{\~a}o da Ortografia de L{\'\i}nguas Mo{\c{c}}ambicanas: Relat{\'o}rio do VI Semin{\'a}rio}.
\newblock Centro de Estudos das L{\'\i}nguas Mo{\c{c}}ambicanas.

\bibitem[{{NLLB Team} et~al.(2022){NLLB Team}, Costa-jussà, Cross, Çelebi, Elbayad, Heafield, Heffernan, Kalbassi, Lam, Licht, Maillard, Sun, Wang, Wenzek, Youngblood, Akula, Barrault, Mejia-Gonzalez, Hansanti, Hoffman, Jarrett, Sadagopan, Rowe, Spruit, Tran, Andrews, Ayan, Bhosale, Edunov, Fan, Gao, Goswami, Guzmán, Koehn, Mourachko, Ropers, Saleem, Schwenk, and Wang}]{nllb-22}
{NLLB Team}, Marta~R. Costa-jussà, James Cross, Onur Çelebi, Maha Elbayad, Kenneth Heafield, Kevin Heffernan, Elahe Kalbassi, Janice Lam, Daniel Licht, Jean Maillard, Anna Sun, Skyler Wang, Guillaume Wenzek, Al~Youngblood, Bapi Akula, Loic Barrault, Gabriel Mejia-Gonzalez, Prangthip Hansanti, John Hoffman, Semarley Jarrett, Kaushik~Ram Sadagopan, Dirk Rowe, Shannon Spruit, Chau Tran, Pierre Andrews, Necip~Fazil Ayan, Shruti Bhosale, Sergey Edunov, Angela Fan, Cynthia Gao, Vedanuj Goswami, Francisco Guzmán, Philipp Koehn, Alexandre Mourachko, Christophe Ropers, Safiyyah Saleem, Holger Schwenk, and Jeff Wang. 2022.
\newblock \href {http://arxiv.org/abs/arXiv:1902.01382} {No language left behind: Scaling human-centered machine translation}.

\bibitem[{NLLBTeam et~al.(2024)NLLBTeam, Costa-jussà, Cross, Çelebi, Elbayad, Heafield, Heffernan, Kalbassi, Lam, Licht, Maillard, Sun, Wang, Wenzek, Youngblood, Akula, Barrault, Gonzalez, Hansanti, Hoffman, Jarrett, Sadagopan, Rowe, Spruit, Tran, Andrews, Ayan, Bhosale, Edunov, Fan, Gao, Goswami, Guzmán, Koehn, Mourachko, Ropers, Saleem, Schwenk, and Wang}]{Costa-jussa2024}
NLLBTeam, Marta~R. Costa-jussà, James Cross, Onur Çelebi, Maha Elbayad, Kenneth Heafield, Kevin Heffernan, Elahe Kalbassi, Janice Lam, Daniel Licht, Jean Maillard, Anna Sun, Skyler Wang, Guillaume Wenzek, Al~Youngblood, Bapi Akula, Loic Barrault, Gabriel~Mejia Gonzalez, Prangthip Hansanti, John Hoffman, Semarley Jarrett, Kaushik~Ram Sadagopan, Dirk Rowe, Shannon Spruit, Chau Tran, Pierre Andrews, Necip~Fazil Ayan, Shruti Bhosale, Sergey Edunov, Angela Fan, Cynthia Gao, Vedanuj Goswami, Francisco Guzmán, Philipp Koehn, Alexandre Mourachko, Christophe Ropers, Safiyyah Saleem, Holger Schwenk, and Jeff Wang. 2024.
\newblock \href {https://doi.org/10.1038/s41586-024-07335-x} {Scaling neural machine translation to 200 languages}.
\newblock \emph{Nature}.

\bibitem[{Papineni et~al.(2002)Papineni, Roukos, Ward, and Zhu}]{papineni-etal-2002-bleu}
Kishore Papineni, Salim Roukos, Todd Ward, and Wei-Jing Zhu. 2002.
\newblock \href {https://doi.org/10.3115/1073083.1073135} {{B}leu: a method for automatic evaluation of machine translation}.
\newblock In \emph{Proceedings of the 40th Annual Meeting of the Association for Computational Linguistics}, pages 311--318, Philadelphia, Pennsylvania, USA. Association for Computational Linguistics.

\bibitem[{Popovi{\'c}(2015)}]{popovic-2015-chrf}
Maja Popovi{\'c}. 2015.
\newblock \href {https://doi.org/10.18653/v1/W15-3049} {chr{F}: character n-gram {F}-score for automatic {MT} evaluation}.
\newblock In \emph{Proceedings of the Tenth Workshop on Statistical Machine Translation}, pages 392--395, Lisbon, Portugal. Association for Computational Linguistics.

\bibitem[{Post(2018)}]{post-2018-call}
Matt Post. 2018.
\newblock \href {https://doi.org/10.18653/v1/W18-6319} {A call for clarity in reporting {BLEU} scores}.
\newblock In \emph{Proceedings of the Third Conference on Machine Translation: Research Papers}, pages 186--191, Brussels, Belgium. Association for Computational Linguistics.

\bibitem[{Vaswani et~al.(2017)Vaswani, Shazeer, Parmar, Uszkoreit, Jones, Gomez, Kaiser, and Polosukhin}]{Vaswani2017}
Ashish Vaswani, Noam Shazeer, Niki Parmar, Jakob Uszkoreit, Llion Jones, Aidan~N Gomez, \L~ukasz Kaiser, and Illia Polosukhin. 2017.
\newblock \href {https://proceedings.neurips.cc/paper/2017/file/3f5ee243547dee91fbd053c1c4a845aa-Paper.pdf} {Attention is all you need}.
\newblock In \emph{Advances in Neural Information Processing Systems}, volume~30. Curran Associates, Inc.

\bibitem[{Wang et~al.(2024)Wang, Adelani, Agrawal, Masiak, Rei, Briakou, Carpuat, He, Bourhim, Bukula, Mohamed, Olatoye, Adewumi, Mokayed, Mwase, Kimotho, Yuehgoh, Aremu, Ojo, Muhammad, Osei, Omotayo, Chukwuneke, Ogayo, Hourrane, El~Anigri, Ndolela, Mangwana, Mohamed, Ayinde, Awoyomi, Alkhaled, Al-azzawi, Etori, Ochieng, Siro, Kiragu, Muchiri, Kimotho, Sakayo, Wamba, Abolade, Ajao, Shode, Macharm, Iro, Abdullahi, Moore, Opoku, Akinjobi, Afolabi, Obiefuna, Ogbu, Ochieng{'}, Otiende, Mbonu, Lu, and Stenetorp}]{wang-etal-2024-afrimte}
Jiayi Wang, David Adelani, Sweta Agrawal, Marek Masiak, Ricardo Rei, Eleftheria Briakou, Marine Carpuat, Xuanli He, Sofia Bourhim, Andiswa Bukula, Muhidin Mohamed, Temitayo Olatoye, Tosin Adewumi, Hamam Mokayed, Christine Mwase, Wangui Kimotho, Foutse Yuehgoh, Anuoluwapo Aremu, Jessica Ojo, Shamsuddeen Muhammad, Salomey Osei, Abdul-Hakeem Omotayo, Chiamaka Chukwuneke, Perez Ogayo, Oumaima Hourrane, Salma El~Anigri, Lolwethu Ndolela, Thabiso Mangwana, Shafie Mohamed, Hassan Ayinde, Oluwabusayo Awoyomi, Lama Alkhaled, Sana Al-azzawi, Naome Etori, Millicent Ochieng, Clemencia Siro, Njoroge Kiragu, Eric Muchiri, Wangari Kimotho, Toadoum~Sari Sakayo, Lyse~Naomi Wamba, Daud Abolade, Simbiat Ajao, Iyanuoluwa Shode, Ricky Macharm, Ruqayya Iro, Saheed Abdullahi, Stephen Moore, Bernard Opoku, Zainab Akinjobi, Abeeb Afolabi, Nnaemeka Obiefuna, Onyekachi Ogbu, Sam Ochieng{'}, Verrah Otiende, Chinedu Mbonu, Yao Lu, and Pontus Stenetorp. 2024.
\newblock \href {https://doi.org/10.18653/v1/2024.naacl-long.334} {{A}fri{MTE} and {A}fri{COMET}: Enhancing {COMET} to embrace under-resourced {A}frican languages}.
\newblock In \emph{Proceedings of the 2024 Conference of the North American Chapter of the Association for Computational Linguistics: Human Language Technologies (Volume 1: Long Papers)}, pages 5997--6023, Mexico City, Mexico. Association for Computational Linguistics.

\bibitem[{Xue et~al.(2022)Xue, Barua, Constant, Al-Rfou, Narang, Kale, Roberts, and Raffel}]{xue-etal-2022-byt5}
Linting Xue, Aditya Barua, Noah Constant, Rami Al-Rfou, Sharan Narang, Mihir Kale, Adam Roberts, and Colin Raffel. 2022.
\newblock \href {https://doi.org/10.1162/tacl_a_00461} {{B}y{T}5: Towards a token-free future with pre-trained byte-to-byte models}.
\newblock \emph{Transactions of the Association for Computational Linguistics}, 10:291--306.

\bibitem[{Xue et~al.(2021)Xue, Constant, Roberts, Kale, Al-Rfou, Siddhant, Barua, and Raffel}]{xue-etal-2021-mt5}
Linting Xue, Noah Constant, Adam Roberts, Mihir Kale, Rami Al-Rfou, Aditya Siddhant, Aditya Barua, and Colin Raffel. 2021.
\newblock \href {https://doi.org/10.18653/v1/2021.naacl-main.41} {m{T}5: A massively multilingual pre-trained text-to-text transformer}.
\newblock In \emph{Proceedings of the 2021 Conference of the North American Chapter of the Association for Computational Linguistics: Human Language Technologies}, pages 483--498, Online. Association for Computational Linguistics.

\end{thebibliography}

\appendix


\begin{table*}[h]
\centering

\begin{tabular}{llp{3cm}p{2cm}l}
\toprule
\textbf{Name} & \textbf{Role}    & \textbf{Tasks} & \textbf{Expertise} & \textbf{Alias}\\
\midrule
Araibo Suhamihe         & Translator & Translate \textit{devtest}, Revise \textit{dev} & Professional experience & Translator1\\ \hline
Salustiano Eurico Ramos & Translator & Translate \textit{dev}, Revise \textit{devtest} & Professional experience & Translator2  \\ \hline
Gito Anastácio Anastácio & Evaluator & Evaluate and post-edit \textit{devtest} / \textit{dev} & Professional experience & Annotator1\\ \hline
Júlio José Paulo        & Evaluator & Evaluate and post-edit \textit{devtest} / \textit{dev} & Professional experience & Annotator2\\ \hline
Vasco André António     & Evaluator & Evaluate and post-edit \textit{devtest} / \textit{dev} & Professional experience & Annotator3\\
                        
\bottomrule
\end{tabular}

\caption{Translation Team }
\label{tab:team}
\end{table*}


\begin{table*}[h]
\centering
\resizebox{0.9\textwidth}{!}{%
\begin{tabular}{lrl}
 & &  \\
\rowcolor{gray!20} \textit{pt} $\to$ \textit{vmw} & &  \\
\toprule
\textbf{Source} & \textit{pt} & \textbf{A camada é mais fina debaixo dos mares e mais espessa abaixo das montanhas}. \\
Translation & \textit{en} & \textit{It is thinner under the maria and thicker under the highlands.} \\
\cmidrule(rl){2-3}
\multirow{2}{*}{\textbf{References (\textit{vmw})}} & A & Mpattapatthaaya tiwoyeva vathi wa mphareya ni yowoneya vathi wa miyaako. \\
& B & Mpattapatthaaya ti'yottettheeya othi wa iphareya ni yookhoomala vathi wa miyaako. \\
\midrule
\multirow{8}{*}{\textbf{Systems (\textit{vmw})}}  & \textbf{baseline} & Khalai atthu yahikhotta vathi-va, khukelela vasulu vaya. \\
& \textbf{afri-byT5} & Okhala wira okathi wa okathi ole ti wootepexa ottuli wa iphareya ni otepexa ottuli wa miyako. \\
& \textbf{afri-mT5} & Nthowa nenlo ninkhala ntoko nsuwa ntoko nsuwa ni ninkhala ntoko nsuwa ni ninkhala ntoko nsuwa ni ninkhala ntoko nsuwa. \\
& \textbf{ByT5} & Ekamada eyo yootepa omalela vathi va iphareya ni yootepa omalela vathi va miyaako. \\
& \textbf{MT0} & Okhala wira ematta eyo enniphwanyaneya ottuli wa maasi, nto ematta eyo enniphwanyaneya ottuli wa maasi. \\
& \textbf{mT5} & Ekatana eyo ti yootepa otthuneya ovikana maasi ni yootepa otthuneya ovikana maasi. \\
& \textbf{M2M100} & Ekaaxa ele ti yootepa otthuneya vathi va ephareya ni yootepa otthuneya vathi va mwaako. \\
& \textbf{NLLB} & Mukattelo ti woorekama vathi vathi wa ophareya ni wootepa maasi vathi wa miyaako.\\

\bottomrule
\end{tabular}%
}

\resizebox{0.9\textwidth}{!}{%
\begin{tabular}{lrl}
\rowcolor{gray!20} \textit{vmw} $\to$ \textit{pt} & &  \\
\toprule
\textbf{Source} & \textit{vmw} & \textbf{Mpattapatthaaya tiwoyeva vathi wa mphareya ni yowoneya vathi wa miyaako.} \\
\cmidrule(rl){2-3}
\textbf{References} &  \textit{pt} & A camada é mais fina debaixo dos mares e mais espessa abaixo das montanhas. \\
\midrule
\multirow{8}{*}{\textbf{Systems (\textit{pt})}}  & \textbf{baseline} & A sua <unk> ainda é a propriedade que existe no <unk> sistema de coisas <unk>. \\
& \textbf{afri-byT5} & A sua aliança é pequena sobre o mar e visível das montanhas. \\
& \textbf{afri-mT5} & A sua vantagem é pequena sobre o mar e pequena sobre os oceanos. \\
& \textbf{ByT5} & O amigo é pequeno sobre o mar e visível sobre os montes.\\
& \textbf{MT0} & O companheiro é pequeno na água e pequeno na água. \\
& \textbf{mT5} & O seu amigo é pequeno na água e pequeno na água. \\
& \textbf{M2M100} & A arca é pequena debaixo do mar e visível debaixo das montanhas. \\
& \textbf{NLLB} & A bacia é barata no fundo do mar e muito clara no fundo das margens. \\

\bottomrule
\end{tabular}%
}

\resizebox{0.9\textwidth}{!}{%
\begin{tabular}{lrl}
 & &  \\
 & &  \\
 & &  \\
 & &  \\
\rowcolor{gray!20} \textit{pt} $\to$ \textit{vmw} & &  \\
\toprule
\textbf{Source} & \textit{pt} & \textbf{Todos os cidadãos da cidade do Vaticano são católicos romanos.} \\
Translation & \textit{en} & \textit{All citizens of Vatican City are Roman Catholic.} \\
\cmidrule(rl){2-3}
\multirow{2}{*}{\textbf{References (\textit{vmw})}} & A & Atthu ootheene opooma wo Vatikaanu anatiini a ekirixitawu ya katolika. \\
& B & Atthu ootheene opooma ya oVatikaanu anatiini a ekirixitawu katolika. \\
\midrule
\multirow{8}{*}{\textbf{Systems (\textit{vmw})}}  & \textbf{baseline} & Anammuttettheni otheene a epooma ya Vatoolika aari aRoma. \\
& \textbf{afri-byT5} & Atthu otheene a epooma ya oVaticano ti makatooliku a oRoma. \\
& \textbf{afri-mT5} & Otheene a epooma ya Vatikaano ti maKatoliko a oRoma. Otheene atthu otheene a epooma ya Vatikaano ti maKatoliko romano. \\
& \textbf{ByT5} & Atthu otheene a epooma ya oVatikano ti makatooliku a oRoma. \\
& \textbf{MT0} & Atthu otheene a epooma ya oVaticano ti maKatolika a oRoma. \\
& \textbf{mT5} & Atthu otheene a epooma ya oVaticano ari maKristau a oRoma. \\
& \textbf{M2M100} & Atthu otheene a epooma ya oVaticano ari maKatoolika a oRoma. \\
& \textbf{NLLB} & Atthu ootheene anikhala epooma ya Vatikaano ti makatooliku a orooma. \\

\bottomrule
\end{tabular}%
}

\resizebox{0.9\textwidth}{!}{%
\begin{tabular}{lrl}
\rowcolor{gray!20} \textit{vmw} $\to$ \textit{pt} & &  \\
\toprule
\textbf{Source} & \textit{vmw} & \textbf{Atthu ootheene opooma wo Vatikaanu anatiini a ekirixitawu ya katolika.} \\
\cmidrule(rl){2-3}
\textbf{References} &  \textit{pt} & Todos os cidadãos da cidade do Vaticano são católicos romanos. \\
\midrule
\multirow{8}{*}{\textbf{Systems (\textit{pt})}}  & \textbf{baseline} & Todos na cidade do Vaticano apela a terra de <unk>. \\
& \textbf{afri-byT5} & Toda a população na cidade do Vaticano realiza a religião católica. \\
& \textbf{afri-mT5} & Todos os cidadãos em Vaticano são religiosos da igreja católica. \\
& \textbf{ByT5} & Toda a população na cidade do Vaticano é religiosa da cristã católica. \\
& \textbf{MT0} & Todos os cidadãos da cidade de Vaticane são cristãos da igreja católica. \\
& \textbf{mT5} & Todos os cidadãos na cidade de Vaticano são cristãos católicos. \\
& \textbf{M2M100} & Todos os cidadãos do Vaticano são cristãos católicos. \\
& \textbf{NLLB} & Todos na cidade do Vaticano são religiosos católicos. \\

\bottomrule
\end{tabular}%
}

\caption{Example of source-reference sentences pairs from \textit{devtest} and outputs from translating source text using models discussed in Section~\ref{sec:setup}}
\label{tab:translations}
\end{table*}


\begin{figure*}[h]
  \centering
  \includegraphics[width=1\textwidth]{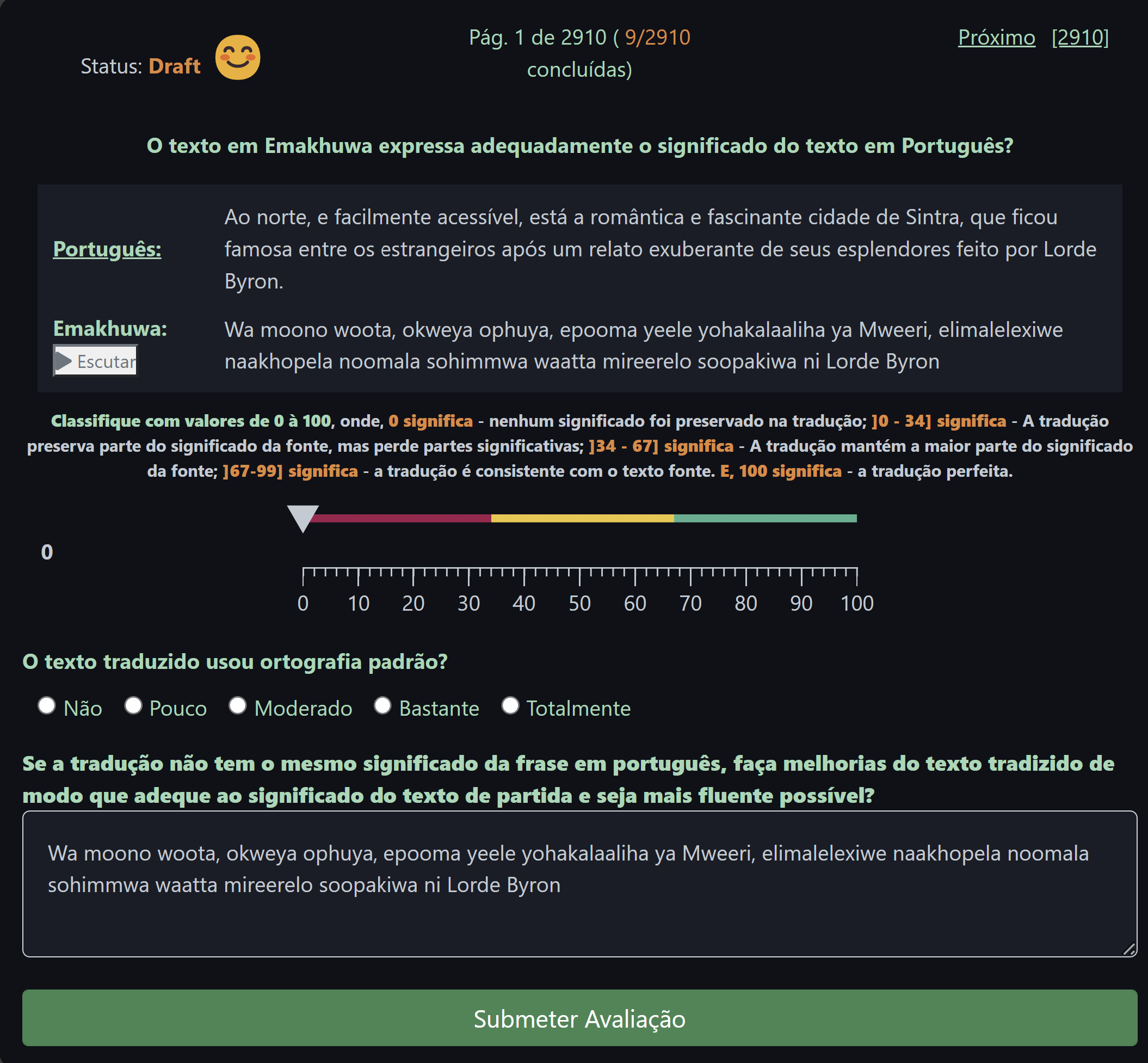}
  \caption{Annotation Tool User Interface.}
  \label{fig:annotation-tool}
\end{figure*}

\begin{figure*}[h]
  \centering
  \includegraphics[width=1\textwidth]{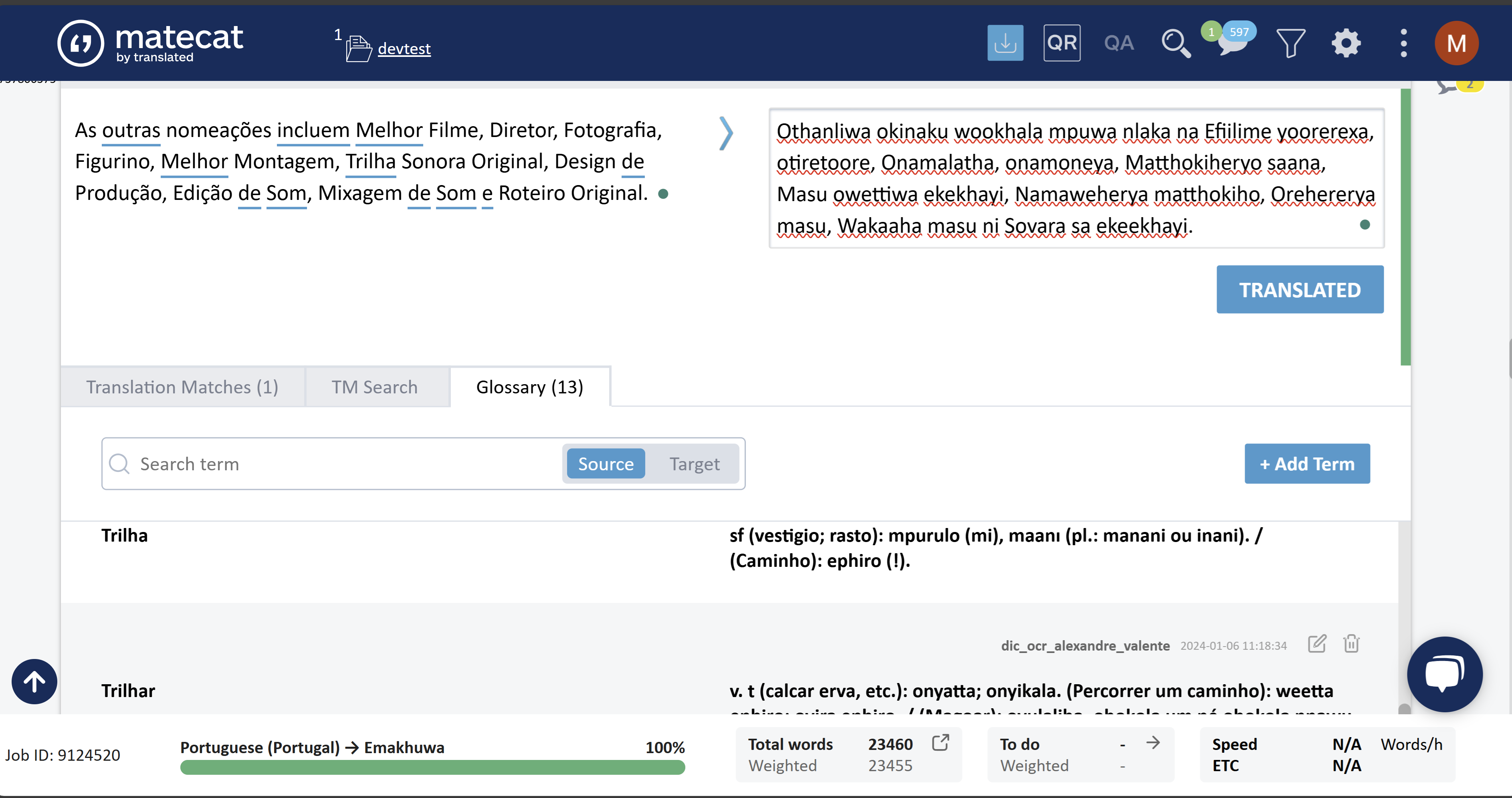}
  \caption{Matecat User Interface}
  \label{fig:matecat-ui}
\end{figure*}

\begin{figure*}[h]
  \centering
  \includegraphics[width=1\textwidth]{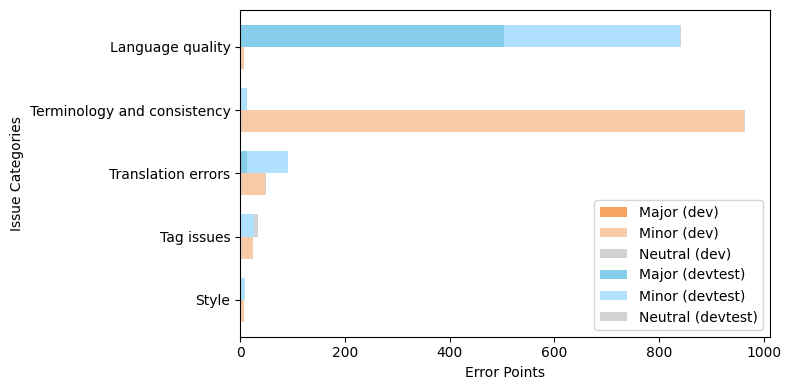}
  \caption{Matecat Quality Report after revision, categorized by the following translation issue typologies: 1) Style (readability, consistent style, and tone); 2) Tag issues (mismatches, whitespaces); 3) Translation errors (mistranslation, additions or omissions); 4) Terminology and translation consistency; 5) Language quality (grammar, punctuation, spelling). The error point count corresponds to the number of segments found with any of the issues described above.}
  \label{fig:errors-points}
\end{figure*}

\begin{figure*}[h]
  \centering
  \includegraphics[width=1\textwidth]{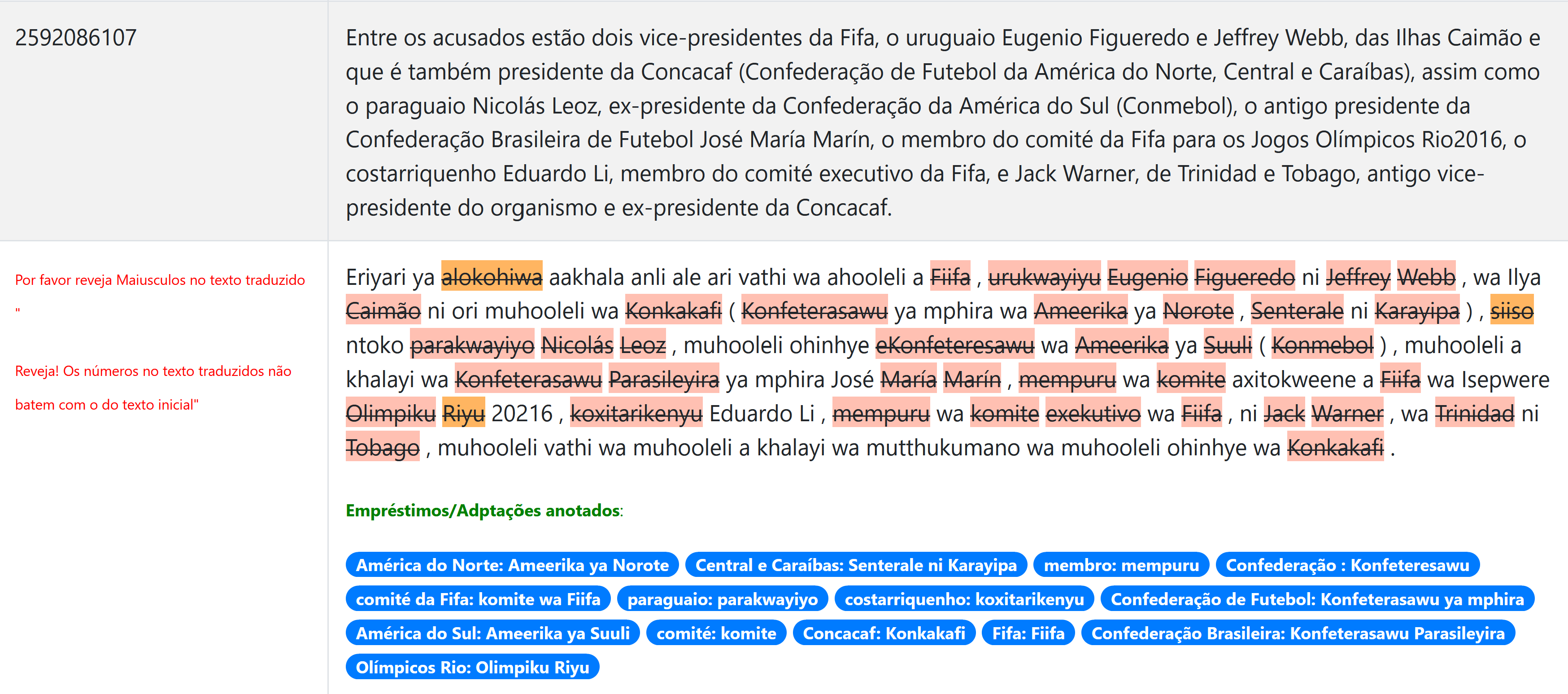}
  \caption{Screenshot of a spelling report. The report is organized into two columns: the first column lists the segment ID along with any potential translation issues (i.e., punctuations, source-target length ratio flag, number mismatch, loanwords not annotated, case mismatch, etc.). The second column displays the source text and its translation. Potential misspellings are highlighted within the translation. In the translation, potential misspellings are highlighted in yellow and red—yellow indicating that suggestions for corrections are available and red indicating that no suggestions exist. Additionally, the report lists all words that translators have annotated as loanwords from Portuguese, using the format \textit{<donor sequence in Portuguese>:<recipient sequence in Emakhuwa>} }
  \label{fig:spelling-report}
\end{figure*}

\end{document}